\documentclass[runningheads]{llncs}
\usepackage[T1]{fontenc}
\usepackage{graphicx}
\usepackage{amsfonts} 
\usepackage{amsmath}
\usepackage{amssymb}
\usepackage{multirow} 
\usepackage{makecell}
\usepackage{subfigure} 
\usepackage{booktabs} 
\usepackage{makecell}
\usepackage{wrapfig}
\usepackage[misc]{ifsym}
\begin{document}
\begin{sloppypar}
%
\title{KnowPrefix-Tuning: A Two-Stage Prefix-Tuning Framework for Knowledge-Grounded \\ Dialogue Generation}
\toctitle{KnowPrefix-Tuning: A Two-Stage Prefix-Tuning Framework for Knowledge-Grounded Dialogue Generation}
\titlerunning{KnowPrefix-Tuning}
%

\author{Jiaqi Bai\inst{1,2} \and
Zhao Yan\inst{3} \and
Ze Yang\inst{2} \and
Jian Yang\inst{2} \and
Xinnian Liang\inst{2} \and  \\
Hongcheng Guo \inst{2} \and 
Zhoujun Li (\Letter) \inst{1,2}}
\tocauthor{Jiaqi Bai, Zhao Yan, Jian Yang, Xinnian Liang, Hongcheng Guo, Zhoujun Li}
\authorrunning{Bai et al.}
%
\institute{
School of Cyber Science and Technology, Beihang University \\
\email{\{bjq,lizj\}@buaa.edu.cn}
\and
State Key Lab of Software Development Environment, Beihang University
\\ \email{\{tobey,jiaya,xnliang,hongchengguo\}@buaa.edu.cn}\\
\and
Tencent Cloud AI \\
\email{zhaoyan@tencent.com}
}

%
%
\maketitle              

\begin{abstract}
Existing knowledge-grounded conversation systems generate responses typically in a retrieve-then-generate manner. 
They require a large knowledge base and a strong knowledge retrieval component, which is time- and resource-consuming. 
In this paper, we address the challenge by leveraging the inherent knowledge encoded in the pre-trained language models (PLMs). 
We propose \textbf{K}nowledgeable \textbf{P}refix \textbf{Tuning} (\textbf{KnowPrefix-Tuning}), a two-stage tuning framework, bypassing the retrieval process in a knowledge-grounded conversation system by injecting prior knowledge into the lightweight knowledge prefix. 
The knowledge prefix is a sequence of continuous knowledge-specific vectors that can be learned during training. 
In addition, we propose a novel interactive re-parameterization mechanism that allows the prefix to interact fully with the PLM during the optimization of response generation.
Experimental results demonstrate that \textbf{KnowPrefix-Tuning} outperforms fine-tuning and other lightweight tuning approaches, and performs comparably with strong retrieval-based baselines while being $3\times$ faster during inference\footnote{The code is available at \url{https://github.com/fantast4ever/KnowPrefix-Tuning}}.

\end{abstract}

\section{Introduction}

Open-domain dialogue system suffers from the problem of generating generic and bland responses, degrading the interaction experience of users \cite{ghazvininejad2018knowledge,holtzman2019curious}. 
Recent efforts follow the paradigm of generating the response by augmenting the source of knowledge associated with the dialogue context \cite{komeili2022internet,sun-etal-2022-multimodal}. 
Knowledge-grounded dialogue (KGD) \cite{dinan2018wizard,ghazvininejad2018knowledge}, as one of the milestone tasks in open-domain dialogue, has attracted many research interests in recent years. The growing number of research works has begun to focus on developing an efficient knowledge-grounded dialogue system \cite{kim2019sequential,lian2019learning,zhao2020knowledge,prabhumoye2021focused,li-etal-2022-knowledge,liu2022multi}.

Existing knowledge-grounded dialogue systems are typically retrieval-augmented \cite{lewis2020retrieval,guu2020retrieval,shuster2021retrieval}, where the knowledge is accessed in an explicit manner.
They first employ a knowledge retrieval component to select knowledge pieces that are most relevant to the dialogue context from a large knowledge base.
Then they augment the selected knowledge pieces with the dialogue context to generate knowledgeable responses.
Although the retrieval-augmented approaches have demonstrated remarkable progress on the KGD task, these approaches inevitably consume considerable resources and time to train and store the parameters of the knowledge retriever, in order to ensure the knowledge retrieval ability of a KGD system.


\begin{figure*}[t]
\centering
\includegraphics[width=12cm]{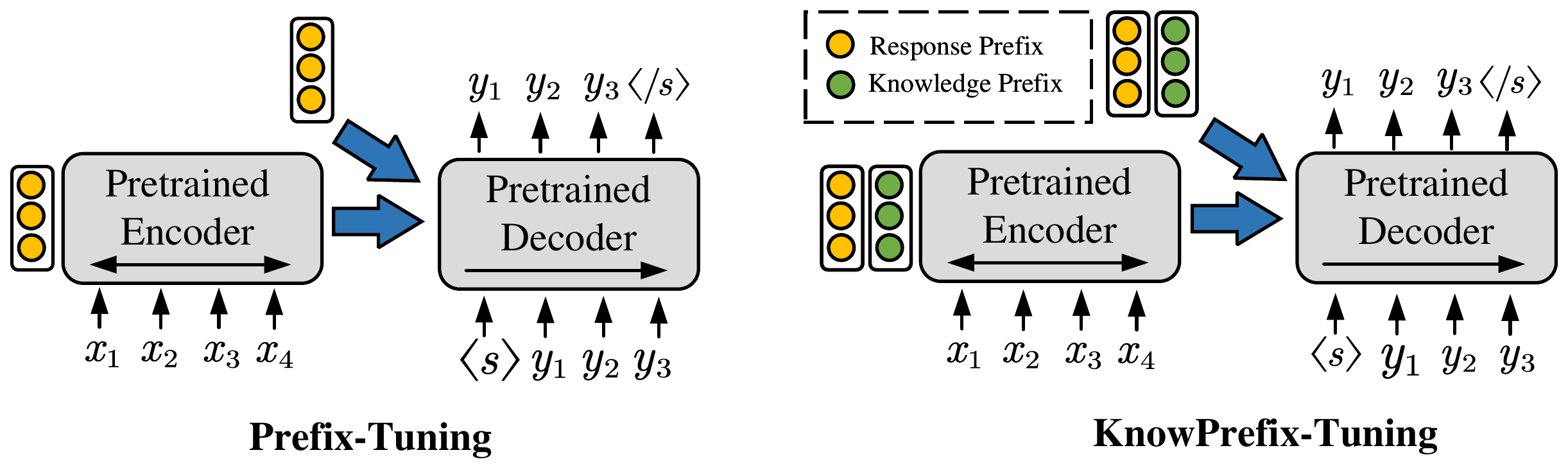}
\caption{Comparison between the prefix-tuning (left part) and the proposed KnowPrefix-Tuning (right part).
Both approaches freeze the pre-trained weights of LM and tune only a small set of parameters that are added as the prefix of the model’s input.
We show the prefix-tuning employing the PLMs with encoder-decoder architecture (e.g., BART). To the PLMs with decoder-only architecture (e.g., GPT2), only the prefix prepended to decoder part is considered.}
\label{fig_method_0}
\vspace{-3mm}
\end{figure*}

Recent studies show that large PLMs carry implicit knowledge \cite{radford2019language,li-etal-2022-knowledge} that can directly apply to downstream generation tasks by proper prompts \cite{brown2020language,li-liang-2021-prefix,karimi-mahabadi-etal-2022-prompt}. 
In particular, Li et al. \cite{li-liang-2021-prefix} proposed prefix-tuning, a lightweight paradigm bypassing finetuning the entire PLM and instead tuning only a small set of parameters that are added as the prefix of the model’s input.
However, the vanilla prefix-tuning approach lacks an effective mechanism to incorporate grounded knowledge (i.e., the knowledge labeled by human annotators) into the generated text. In this paper, we aim to steer the prefix-tuning to the KGD task, where the knowledge can be accessed in an implicit manner to achieve knowledgeable response generation.
There are two key challenges:
(1) Eliciting the knowledge irrelevant to the dialogue context from PLM may mislead the model into generating the context-irrelevant response.
(2) There can be one-to-many relations between the dialogue context and the knowledge to be selected \cite{lian2019learning}. 
Explicitly enumerating all of this knowledge encoded in the PLM is impractical.


To tackle the above challenges, we propose Knowledgeable Prefix-Tuning (\textbf{KnowPrefix-Tuning}),
Figure \ref{fig_method_0} shows the comparison between the prefix-tuning \cite{li-liang-2021-prefix} and the proposed approach.
The proposed KnowPrefix-Tuning is a two-stage prefix-tuning framework.
In the first tuning stage, we inject the prior knowledge into the lightweight knowledge prefix by forcing the model to generate the knowledge grounded on the dialogue context, which facilitates the model to focus more on context-relevant knowledge.
In the second tuning stage, we prompt the PLM to generate the knowledgeable response by response prefix, grounding on the dialogue context and the previously learned knowledge prefix. 
To enhance the interaction between the knowledge and dialogue context, we propose a novel \textbf{Interactive Re-parameterization} mechanism, which further facilitates knowledgeable response generation by encouraging the interaction between the prefix and the PLM.
Experimental results on two knowledge-grounded benchmarks show that the proposed method brings both performance and efficiency improvements. It is $3\times$ faster than the state-of-the-art retrieval-based method during inference stage and outperforms competitive light-weight tuning approaches significantly.

Our contributions are three-fold:
(1) We propose a novel KnowPrefix-Tuning approach for the KGD task. The proposed method bypasses the retrieval process and does not require finetuning the entire PLM. 
(2) We propose a novel interactive re-parameterization mechanism, which allows the interaction between the embedding of the prefix and the PLM during the re-parameterization of the prefix.
(3) We conduct sufficient experiments and qualitative analysis to prove the effectiveness of our proposed methods on two knowledge-grounded dialogue datasets.
\section{Related Work}

\subsection{Knowledge-Grounded Dialogue}
Knowledge-grounded dialogue has shown tremendous potential in enriching and diversifying the responses generated by dialogue agents \cite{meng2020dukenet,shuster2021retrieval,cui2021knowledge,sun-etal-2022-stylized,zhao-etal-2022-learning-express,li2022eliciting}. 
Most of the existing KGD works focus on improving the knowledge retrieval performance, which explicitly retrieves proper knowledge pieces, thereby enhancing the knowledgeable response generation ability \cite{lian2019learning,kim2019sequential,zhao2020knowledge,li2022eliciting}.
Very recent work begins to focus on leveraging the knowledge that is inherently encoded in the PLM \cite{li-etal-2022-knowledge,liu2022multi,xu2022retrieval,DBLP:journals/corr/abs-2212-10400}. 
For example, Li et al. \cite{li-etal-2022-knowledge} employed the pre-train technique to encode the multi-source knowledge into a unified framework. 
Lie et al. proposed to \cite{liu2022multi} retrieve a small collection of dialogue samples to construct prompts, which is used to  guide the knowledgeable response generation in a multi-stage generation manner.
Li et al. \cite{li2022eliciting} employ PLATO-KAG \cite{huang2021plato} as a backbone, using the generated knowledge as a noisy knowledge source and propose the posterior-based reweighing and the noisy training strategy to improve the performance of a knowledge retriever.
Compared with above works, we do not explicitly retrieve any knowledge pieces from the data base. We instead encode the knowledge into the lightweight knowledge prefix, which saves much computational resource and time.

\subsection{Prompting in Language Models}
Pre-trained language models (PLMs) \cite{kenton2019bert,lewis2020bart,mbart} exhibit an innate ability to store commonsense knowledge \cite{davison2019commonsense,zhou2020evaluating,cui2021commonsense,lester2021power}. 
It can be prompted to do many downstream natural language generation task, such as closed-book question answering (QA) \cite{petroni2019language,wang2021can}, text summarization \cite{li-liang-2021-prefix,zhao-etal-2022-domain}, and so on.
Earlier work use human written prompts by manually designing prompt templates \cite{brown2020language,schick2021s}, or search prompts over the discrete space of words \cite{shin2020autoprompt}.
Recent work focus on continuous prompt learning \cite{li-liang-2021-prefix,lester2021power}, where the prompts are represented as a group of vectors that can be optimized during training process.
Our work is enlightened from Prefix-Tuning \cite{li-liang-2021-prefix}. The main difference is that the proposed KnowPrefix-Tuning is a two-stage tuning framework that encodes the global knowledge into a group of continuous knowledge-specific vectors, allowing the interaction between the embedding of prefixes and the PLM.

\section{Background}

\subsection{Problem Formalization}

Suppose we have a $T$-turn conversation
$\mathcal{C}=(U_i,K_i,Y_i)_{i=1}^{T}$
in a knowledge-grounded conversation dataset $\mathcal{D}$, where 
$\forall i$, $(U_i, K_i, Y_i)$ 
is a triplet of query-knowledge-response at turn $i$.
Given an input sequence 
$X_i=( Y_{i-1},U_i )$,
our goal is to generate a knowledgeable response $\widetilde{Y}_i$ by learning a response generation model
$P(Y_i,K_i|X_i;\varTheta )$. 
Existing retrieval-augmented methods tackle this problem by firstly retrieving related knowledge $\widetilde{K}_i$, then augmenting it with input sequence $X_i$.
Here, we propose to bypass the retrieval process by injecting the knowledge $K_i$ into the model parameters $\varTheta$.
Thus, the response can be generated sorely based on the input sequence $X_i$.

\subsection{Prefix-Tuning}
\label{backbone}
In this section, we briefly describe Prefix-Tuning \cite{li-liang-2021-prefix} upon the vanilla Transformer architecture \cite{vaswani2017attention}, based on which we can validate the effectiveness of our approach.


In the vanilla Transformer architecture, each transformer layer equips with multiple attention heads, and each head attends over the tokens of the input context. 
Let $\mathbf{H}_l$ denote the output of a single attention head in the $l$-th Transformer layer, which is formalized by:
\begin{equation}
\label{hid_states}
\mathbf{H}_l=\mathcal{F}_{\varLambda}\left( Q_l\mathbf{W}_{l}^{Q},K_l\mathbf{W}_{l}^{K},V_l\mathbf{W}_{l}^{V} \right) \in \mathbb{R}^{N\times d}
\end{equation}

\noindent where $\mathcal{F}( \cdot )$ denotes the attention computational function, $\varLambda$ denotes an attention type. 
$\varLambda \in \{ E_S,D_C,D_S \}$ where $E_S$ is the encoder self-attention, $D_C$ is the decoder cross-attention and $D_S$ is the decoder masked self-attention, respectively.
$Q_l\in \mathbb{R}^{N\times d}$
is the query matrix in the 
$l$-th layer. 
$K_l,V_l\in \mathbb{R}^{M\times d}$ 
denote the $l$-th layer key and value matrix,
where $N$ is the sequence length related to queries, $M$ is the sequence length related to key and value. 
$\mathbf{W}_{l}^{Q}$, $\mathbf{W}_{l}^{K}$, $\mathbf{W}_{l}^{V}\in \mathbb{R}^{d\times d}$ are head-specific projection weights for $Q$, $K$ and $V$. respectively.

In prefix-tuning, the prefix is denoted as a distinct key-value pair for the attention type $\varLambda$, which is a set of continuous specific vectors that can be learned by
\begin{equation}
\label{prefix_repar}
P_{\varLambda}=\mathrm{MLP}\left( \mathbf{E}\left( X_{\varLambda} \right) \right) \in \mathbb{R}^{2\times \rho Ld}
\end{equation}

\noindent where $X_{\varLambda}\in \mathbb{R}^{\rho}$ is the input token of prefix, $\mathbf{E}( \cdot )$ is the embedding projection matrix. 
$\rho$ is the length of prefix token. 
$P_{\varLambda}=\{P_{\varLambda}^{( 1 )},\cdots P_{\varLambda}^{( L )}\}$ denotes a prefix set for $\varLambda$, where $L$ is the number of layer in transformer.
$\forall l\in \{ 1,\cdots, L \}$, 
$P_{\varLambda}^{( l )}=( P_{\varLambda ,K}^{( l )},P_{\varLambda ,V}^{( l )} ) \in \mathbb{R}^{2\times \rho d}$, 
$d$ is the embedding dimension. 
$P_{\varLambda ,K}^{( l )}$ 
and 
$P_{\varLambda ,V}^{( l )}$ 
are the key and value of prefix in the $l$-th layer, respectively.
During the prefix-tuning stage, the key-value pair in equation \ref{hid_states} is augmented to become
\begin{align}
K_l &\gets \left[ P_{\varLambda ,K}^{(l)};K_l \right] \in \mathbb{R}^{\left( M+\rho \right) \times d} \\
V_l &\gets \left[ P_{\varLambda ,V}^{(l)};V_l \right] \in \mathbb{R}^{\left( M+\rho \right) \times d}
\end{align}
\section{Approach}



\subsection{KnowPrefix-Tuning}

Formally, given the dialogue context 
$X=\{ x_1,x_2,\cdots x_{| X |} \}$.
The corresponding knowledge piece is
$K=\{ k_1,k_2,\cdots k_{| K |} \}$
and the response 
$Y=\{ y_1,y_2,\cdots y_{| Y |} \}$.
In the \textbf{first} tuning stage, we obtain the \emph{knowledge prefixes} by applying the vanilla prefix-tuning approach, feeding the dialogue context $X$ to the model and asking it to predict the knowledge $K$ token by token, 
which can be realized by optimizing following loss:
\begin{equation}
\label{know_prefix_gen}
\mathcal{L}_1=-\mathbb{E}_{\left( X,K \right) \in \mathcal{D}}\sum_{t=1}^{\left| K \right|}{\log P\left( \left. k_t \right|X,k_{1:\left( t-1 \right)};\theta _{LM},\theta _K \right)}
\end{equation}

\noindent where $\theta _{LM}$ are the parameters of PLM, $\theta _{LM}$ are holding fixed during the tuning procedure. 
$\theta _K$ are the learnable parameters for the knowledge prefix. 

In the \textbf{second} tuning stage, we fix the parameters during optimizing the knowledge prefix in the first tuning stage, and add additional learnable \emph{response prefixes} to guide the knowledgeable response generation.
Formally, Given the dialogue context $X$, our goal is to generate the response $Y$ one token at a time. 
The generation process of $Y$ can be defined by optimizing the following loss:
\begin{equation}
\mathcal{L}_2=-\mathbb{E}_{\left( X,Y \right) \in \mathcal{D}}\sum_{t=1}^{\left| Y \right|}{\log P\left( \left. y_t \right|X,y_{1:\left( t-1 \right)};\theta _{LM},\theta _K,\theta _Y \right)}
\end{equation}

\noindent where $\theta _Y$ are the parameters for knowledgeable response generation. During the second tuning stage, the parameters of $\theta _{LM}$ and $\theta _K$ are holding fixed, only the parameters of $\theta _Y$ are updated.


\subsection{KnowPrefix-Tuning with Interactive Re-parameterization}

Re-parameterization has been demonstrated to be indispensable in boosting the performance of prefix-tuning \cite{li-liang-2021-prefix}. It is realized by introducing a large feed-forward neural network during the optimization of the prefix, as we introduced in equation \ref{prefix_repar}. 
While for the KGD task, the interaction between the knowledge and the dialogue context is still significant to realize the knowledgeable response generation.
To model this kind of interaction, we propose the interactive re-parameterization mechanism, which considers the embedding of prefix and PLM as interfaces, and conducting the multi-head attention \cite{vaswani2017attention} to realize the interaction between them.




To re-parameterize the prefix $P_{\varLambda}$, we concatenate an interaction term $\mathbf{I}^o$ with the embedding of prefix $\mathbf{E}_{\theta _Y}( X_{\varLambda})$, and re-parameterize the concatenation between them, which is given by:
\begin{equation}
P_{\varLambda}=f_{\theta _{\varLambda}}\left( \left[ \mathbf{E}_{\theta _Y}\left( X_{\varLambda} \right) ;\mathbf{I}^o \right] \right) \in \mathbb{R}^{2\times \rho Ld}
\end{equation}

\noindent where $\left[ \cdot ;\cdot \right]$ denotes the concatenation operation. $f_{\theta _{\varLambda}}\left( \cdot \right)$ is an interactive function, which can be any neural network such as a MLP.
We define the interaction term $\mathbf{I}^o$ as a weighted sum for the embedding of PLM, which can be obtained by:
\begin{align}
&\mathbf{I}^o=\mathbf{I}^e\mathbf{E}_{\theta _{LM}}\in \mathbb{R}^{\rho \times d} \\
&\mathbf{I}^e=\mathrm{softmax} \left( \mathbf{H}^o\mathbf{E}_{\theta _{LM}}^{T} \right) \in \mathbb{R}^{\rho \times \left| V_{LM} \right|}
\label{interaction_term}
\end{align}

\noindent where $\mathbf{E}_{\theta _{LM}}\in \mathbb{R}^{| V_{LM} |\times d}$ is the embedding matrix of the PLM and $| V_{LM} |$ is its vocabulary size. 
$\mathbf{H}^o$ is the output state for each token of prefix sequence. 
$\mathbf{I}^e$ measures the contextual similarity between the $\mathbf{H}^o$ and the embedding of language model $\mathbf{E}_{\theta _{LM}}$.

We obtain the output state $\mathbf{H}^o$ by considering the interaction of embedding between knowledge prefixes and response prefixes, which can be realized by the following:
\begin{gather}
\mathbf{H}^o=\left[ \mathbf{H}_1;\cdots ;\mathbf{H}_N \right] \mathbf{W}^o
\\
\mathbf{H}_j=\mathrm{Attention}\left( Q\mathbf{W}_{Q1}^{j},K\mathbf{W}_{K1}^{j},V\mathbf{W}_{V1}^{j} \right) 
\\
Q=\mathrm{Tanh}\left( \mathbf{E}_{\theta _Y}\left( X_{\varLambda}^{Y} \right) \mathbf{W}_{Q2}^{j} \right) 
\\
K=\mathrm{Tanh}\left( \mathbf{E}_{\theta _K}\left( X_{\varLambda}^{K} \right) \mathbf{W}_{K2}^{j} \right) 
\\
V=\mathrm{Tanh}\left( \mathbf{E}_{\theta _K}\left( X_{\varLambda}^{K} \right) \mathbf{W}_{V2}^{j} \right) 
\end{gather}

\noindent  $N$ is the number of attention heads.
$\mathbf{W}^o\in \mathbb{R}^{d_m\times d}$ is the projection weights of the concatenated output for all attention heads. 
$d_m$ is the dimension of the hidden states in the re-parameterization network. 
$\mathbf{H}_j\in \mathbb{R}^{d_h\times d}$ is the output state for $j$-th attention head.
$d_h=d_m//N$ is the output dimension for each attention head.
$\mathbf{W}_{Q}^{j}$, $\mathbf{W}_{K}^{j}$ and $\mathbf{W}_{V}^{j}$ are head-specific projections for $Q$, $K$ and $V$, respectively. 

To enforce the re-parameterization module eliciting the proper knowledge from the embedding of PLM, inspired by Zhao et al. \cite{zhao2017learning} and Bao et al. \cite{bao-etal-2021-plato}, we supervise the term $\mathbf{I}^e$ by an additional loss $\mathcal{L}_{bow}$ with the bag of words $B_Y$ in the ground-truth response, where the bag of words are obtained by removing the punctuation and the stopwords in response $Y$, forcing the model focus more on content words.
The $\mathcal{L}_{bow}$ can be obtained by:
\begin{equation}
\mathcal{L}_{bow}=-\mathbb{E}_{\mathbf{I}_{w}^{e}\sim P\left( \left. \mathbf{I}^e \right|X,Y \right)}\frac{1}{\left| B_Y \right|}\sum_{w\in B_Y}{\log}\left( \mathbf{I}_{w}^{e} \right) 
\end{equation}

\noindent Intuitively, the bag of words loss  $\mathcal{L}_{bow}$ discards the words orders and forces the term $\mathbf{I}_{w}$ to capture the global information of the target response.
The overall loss in the second stage is defined by:
\begin{equation}
\label{overall_loss}
\mathcal{L}_2\gets \mathcal{L}_2+\mathcal{L}_{bow}
\end{equation}

\section{Experiments}

\subsection{Datasets and Baseline Models}

\paragraph{Datasets} We conduct our experiment on two commonly used knowledge-grounded dialogue datasets: Wizard of Wikipedia (Wizard) \cite{dinan2018wizard}, and CMU Document Grounded Conversations (CMU\_Dog) \cite{zhou2018dataset}. 
The Wizard and CMU\_DoG datasets are constructed by the Amazon Mechanical Turk workers and employ Wikipedia as the knowledge base.
Wizard is split into 18,430 training dialogues on 1247 topics, 1,948 validation dialogues on 599 topics and 1,933 test dialogues on 591 topics. 
The test set is further split into test seen set and unseen set according to the topics. 
The test seen set contains 965 dialogues on 533 topics. 
The test unseen set contains 968 dialogues on 58 topics whose topics are never seen in the training and validation set. 
There are about 9.0  turns on average in each dialogue of the dataset.
CMU\_DoG is split into 3373 training dialogues, 229 validation dialogues, and 619 test dialogues. 
There are 30 topics in the dataset on average, and each dialogue has about 22.0 turns on average.
We implement the pre-processing for both WoW and CMU\_DoG datasets with the code published on ParlAI{\footnote{\url{https://github.com/facebookresearch/ParlAI}}}.

\paragraph{Baseline Models}
We compare our approach with two types of knowledge access methods: 
The first group is \textbf{Explicit Knowledge Access}, which explicitly retrieves knowledge from a knowledge base.
Then use the retrieved knowledge augmenting with the dialogue context to guide the knowledgeable response generation:
\textbf{i)} Transformer Memory Network (TMN) \cite{dinan2018wizard}: 
The model combines Transformer \cite{vaswani2017attention} with an external memory network in an end-to-end manner, which introduces an additional loss to better select knowledge.
\textbf{ii)} Sequential Knowledge Transformer (SKT) \cite{kim2019sequential}: 
The model employs sequential variable model to conduct knowledge selection for response generation, which considers the interaction between the history of knowledge selection and dialogue context.
\textbf{iii)} ZRKGC
\cite{li2020zero}:
The model employs pre-train techniques to handle the zero-resource challenge in KGD task. We choose the one that uses the full training data version for a fair comparison.
\textbf{iv)} KnowledGPT
\cite{zhao2020knowledge}:
The model employs reinforcement learning to jointly optimize knowledge selection and response generation in a joint manner.
\textbf{v)} PLATO-KAG$^+$
\cite{li2022eliciting}:
The model employs PLATO-KAG \cite{huang2021plato} as a backbone, treating the generated knowledge as a noisy knowledge source and propose the posterior-based reweighing and the noisy training strategy to enhance the knowledge retrieval ability.

The second group contains \textbf{Implicit Knowledge Access} methods, which does not require to explicitly retrieve or generate knowledge to guide the knowledgeable response generation:
\textbf{i)} KnowExpert
\cite{xu2022retrieval}: 
The model employs topic modelling technique to inject prior knowledge into the GPT-2 with lightweight adapters. We report the results of their model under weighted-sum (KnowExpert$_{\mathrm{w}}$) setting.
\textbf{ii)} Fine-Tuning:
We fine-tune the whole model of GPT-2 and BART for response generation.
It is to check if only the general fine-tuning approach works well on this task.
\textbf{iii)} Prefix-Tuning
\cite{li-liang-2021-prefix}: 
We apply the Prefix-Tuning approach to generate responses based on the given dialogue context, without the supervision of the external knowledge base. 
It is to check if only the prefix-tuning approach works well on this task. We employ both BART and GPT-2 as backbone PLMs to conduct response generation.
\textbf{iv)} Knowledgeable Prefix-Tuning (KnowPrefix-Tuning):
The method proposed in our paper. We use both GPT-2 and BART as the backbone to conduct the experiment.

\subsection{Evaluation Metrics}

We conduct both automatic and human evaluations. For automatic evaluation, following the previous work on KGC \cite{dinan2018wizard,shuster2021retrieval,komeili2022internet}, we report perplexity (PPL), F1 and Knowledge F1 (KF1) metrics.
The perplexity (PPL) measures how likely the model can generate human-like responses. 
The F1 score indicates the unigram overlap between the generated response and the reference response.
The KF1 score measures the overlap between the generated response and the knowledge \emph{on which the human grounded during dataset collection} \cite{komeili2022internet}, which captures whether a model is speaking knowledgeably by using the knowledge relevant to dialogue context.
To evaluate how the number of parameters impacts the model performance, following Li et al. \cite{li-liang-2021-prefix}, we count the number of parameters that are fine-tuned in each method.

Apart from the automatic evaluation, we conduct the human evaluation from three aspects of the generated response: 
\textbf{Fluency} measures how fluent the generated responses of the model are. 
\textbf{Context coherency} measures how the generated responses are relevant to the dialogue context. 
\textbf{Knowledge relevancy} measures how knowledgeable the generated responses are, according to the amount of new knowledge introduced into the dialogue and the factuality of the generated response.
We employ three well-educated annotators for human evaluation.  
Concretely, each annotator is shown in an example containing a dialogue context and model responses that are randomly shuffled to hide their sources.
We randomly select 100 examples from the test set (both test seen and test unseen for the Wizard dataset), and ask each annotator to assign a score in \{0, 1, 2\} to each response for each aspect. 
The agreement among the annotators is measured via Fleiss’ kappa \cite{fleiss1971measuring}.

\subsection{Implementation Details}
We choose BART{\footnotesize{LARGE}} (406M), GPT2{\footnotesize{LARGE}} (774M) to adapt our KnowPrefix-Tuning framework\footnote{We implement the model with the code shared in \url{https://github.com/huggingface-/transformers}.}.
In both knowledge prefix-tuning stages, we set the same length of the prefix token to 20. 
The embedding size of the prefix token is set to the same embedding size of PLM. 
The hidden states of the re-parameterization network are set to 800. 
When generating knowledgeable responses, we fix the decoding parameters to beam search (beam size 3) with a minimum sequence length of 20 and beam blocking of 3-grams within the response, similar to choices in \cite{roller2021recipes,komeili2022internet}.
All models are learned with the AdamW \cite{loshchilov2018fixing} optimizer with learning rate 3e-5 in 40 epochs. 
We set the warm-up steps to 2000 and applied a linear learning rate scheduler with a batch size of 32. 
Each experiment is conducted on Tesla V-100 machines.

\begin{table*}[t]
\centering
\caption{Automatic evaluation results on the Wizard and CMU\_DoG datasets. Bold face indicates the best result in terms of the corresponding metric.``\#Para'' denotes the number of fine-tuned parameters in each method. PPL values are not comparable across different backbone PLMs as they use different dictionaries.}
\renewcommand\arraystretch{1.1}
\resizebox{1.0\linewidth}{!}{
\begin{tabular}{l ccc ccc ccc c}
    \hline
    \hline
    
    \multirow{3}*{\textbf{Models}} & 
    \multicolumn{3}{c}{\textbf{Wizard Seen}} & 
    \multicolumn{3}{c}{\textbf{Wizard Unseen}} & 
    \multicolumn{3}{c}{\textbf{CMU\_DoG}} & 
    \multirow{3}*{\textbf{\#Para}} \cr

    \cmidrule(lr){2-4} \cmidrule(lr){5-7} \cmidrule(lr){8-10} &
    
    \textbf{PPL}$\downarrow$ & \textbf{F1}$\uparrow$ &  \textbf{KF1}$\uparrow$ &
    \textbf{PPL}$\downarrow$ & \textbf{F1}$\uparrow$ & \textbf{KF1}$\uparrow$ &
    \textbf{PPL}$\downarrow$ & \textbf{F1}$\uparrow$ & \textbf{KF1}$\uparrow$ \\

    \hline

    \multicolumn{11}{c}{\emph{Explicit Knowledge Access}} \\

    \hline

    TMN & 66.5 & 15.9 & - & 100+ & 14.3 & - & 75.2 & 9.9 & - & $1.6\times 10^7$ \\
    
    SKT & 52.2 & 19.4 & - & 81.5 & 16.2 & - & 42.0 & 9.7 & - & $1.7\times 10^8$ \\

    ZRKGC & 40.4 & 18.7 & - & 41.6 & 18.6 & - & 53.6 & 12.5 & - & $3.3\times 10^8$ \\
    
    KnowledGPT & 19.2 & \textbf{22.0} & - & 22.3 & \textbf{20.5} & - & \textbf{20.0} & \textbf{13.7} & - & $2.4\times 10^8$ \\

    PLATO-KAG$^+$ & \textbf{12.4} & 21.1 & - & \textbf{13.8} & 20.3 & - & - & - & - & $1.6\times 10^{10}$ \\

    \hline

    \multicolumn{11}{c}{\emph{Implicit Knowledge Access}} \\

    \hline

    
    GPT-2 + KnowExpert$_{\mathrm{w}}$ & 15.3 & 18.7 & 14.5 & 20.1 & 16.7 & 12.1 & 17.2 & 12.5 & 4.0 & $1.2\times 10^8$ \\

    GPT-2 + Fine-Tuning & 15.1 & 19.8 & 17.4 & 21.3 & 16.8 & 13.9 & 16.7 & 13.8 & 4.9 & $7.7\times 10^8$ \\

    GPT-2 + Prefix-Tuning & 15.8 & 19.1 & 16.3 & 20.7 & 17.0 & 13.0 & 19.9 & 13.2 & 4.2 & $1.8\times 10^6$ \\

    \textbf{GPT-2 + KnowPrefix-Tuning (Ours)} & 15.2 & 20.1 & \textbf{18.0} & 19.3 & 18.0 & 14.6 & 17.4 & 14.1 & \textbf{5.3} & $3.7\times 10^6$ \\

    \hline

    BART + Fine-Tuning & 14.9 & \textbf{20.3} & 17.8 & 19.1 & 17.3 & 13.8 & \textbf{15.3} & 14.1 & 4.7 & $4.1\times 10^8$ \\

    BART + Prefix-Tuning & 15.2 & 19.2 & 16.5 & 19.4 & 16.9 & 13.4 & 18.0 & 13.9 & 4.2 & $1.5\times 10^6$ \\

    \textbf{BART + KnowPrefix-Tuning (Ours)} & \textbf{14.0} & \textbf{20.3} & 17.4 & \textbf{17.5} & \textbf{18.3} & \textbf{14.9} & 16.3 & \textbf{14.6} & 5.2 & $2.9\times 10^6$ \\


    
    


    \hline
    \hline
\end{tabular}
}
\label{main_results}
\end{table*}

\subsection{Main Results}

\subsubsection{Automatic Evaluation Results}

Table \ref{main_results} reports the automatic evaluation results on Wizard and CMU\_DoG datasets. 
We have the following observations:
(1) Compared with the \emph{Explicit Knowledge Access} methods, KnowPrefix-Tuning achieves competitive performance compared to all of the baselines over both datasets.
Specifically, KnowPrefix-Tuning (BART{\footnotesize{LARGE}}) outperforms retrieval-based baselines on the CMU\_DoG dataset. 
Concretly, the PPL and F1 of KnowPrefix-Tuning (BART{\footnotesize{LARGE}}) outperform the percentage of the strongest baseline KnowledGPT by 19.9\% and 7.0\%, respectively.
The lower PPL indicates that the model prefers to generate more context-relevant responses.
In addition, the KnowPrefix-Tuning (BART{\footnotesize{LARGE}}) has only about 3M parameters that should be fine-tuned. It is only 1\% number of parameters in KnowledGPT updated.
(2) Compared with other \emph{Implicit Knowledge Access} methods, the proposed KnowPrefix-Tuning outperforms the Prefix-Tuning substantially, indicating the effectiveness of our proposed tuning framework. 
In addition, the KnowPrefix-Tuning outperforms Fine-tuning on both Wizard and CMU\_Dog datasets with only 0.7\% parameters updated.
This improvements are more clear on the Wizard Unseen dataset, which indicates that the proposed KnowPrefix-Tuning has a powerful generalization ability to the unseen topics even equipped with fewer parameters. 


\subsubsection{Human Evaluation Results}

Table \ref{human_results} reports the human evaluation results on Wizard dataset. 
We can observe that:
(1) The kappa values are larger than 0.6, indicating substantial agreement among the annotators. 
(2) According to the metric \emph{fluency}, we find that the KnowPrefix-Tuning approach tends to generate more fluent responses.
This result is consistent with the automatic evaluation results in which the KnowPrefix-Tuning shows lower perplexity.
We think it is because that the knowledge prefix learned much context-relevant knowledge through the knowledge generation process, which can provide effective guidance for knowledgeable response generation.
(3) According to the metric \emph{knowledge relevancy}, we observe that both Fine-Tuning and  KnowPrefix-Tuning perform better on the Wizard Seen dataset, while KnowPrefix-Tuning is superior to the Wizard Unseen dataset.
We think it is highly likely that the proposed approach does not disturb the implicit knowledge distribution encoded in PLM during tuning procedure. Thus it can better generalize to the unseen topic during knowledgeable response generation.

\begin{table}[t]
\centering
\caption{Human evaluation results on the Wizard dataset. ``FL'', ``CC'' and ``KR'' are the abbreviation of the ``Fluency'', ``Context Coherency'' and ``Knowledge Relevancy'', respectively.}
\renewcommand\arraystretch{1.1}
\resizebox{0.75\linewidth}{!}{
\begin{tabular}{l cccc cccc}
    \hline
    \hline
    
    \multirow{3}*{\textbf{Models}} & 
    \multicolumn{4}{c}{\textbf{Wizard Seen}} & 
    \multicolumn{4}{c}{\textbf{Wizard Unseen}} \cr 

    \cmidrule(lr){2-5} \cmidrule(lr){6-9} &
    
    \textbf{FL} & \textbf{CC} & \textbf{KR} & \textbf{Kappa} & 
    \textbf{FL} & \textbf{CC} & \textbf{KR} & \textbf{Kappa} \\

    \hline

    KnowExpert$_{\mathrm{w}}$ & 1.87 & 1.62 & 1.55 & 0.65
    & 1.83 & 1.52 & 1.30 & 0.64 \\

    Fine-tuning & 1.89 & \textbf{1.66} & \textbf{1.61} & 0.63 
    & 1.84 & 1.53 & 1.38 & 0.62 \\
    
    Prefix-Tuning & 1.86 & 1.60 & 1.57 & 0.60 
    & 1.82 & 1.52 & 1.35 & 0.61 \\
    
    KnowPrefix-Tuning & \textbf{1.90} & 1.64 & \textbf{1.61} & \textbf{0.65} 
    & \textbf{1.88} & \textbf{1.54} & \textbf{1.43} & \textbf{0.65} \\ 
    
    \hline
    \hline
\end{tabular}
}
\label{human_results}
\end{table}

\subsection{Discussions}

\begin{figure}[t]
\centering
\includegraphics[width=5.5cm]{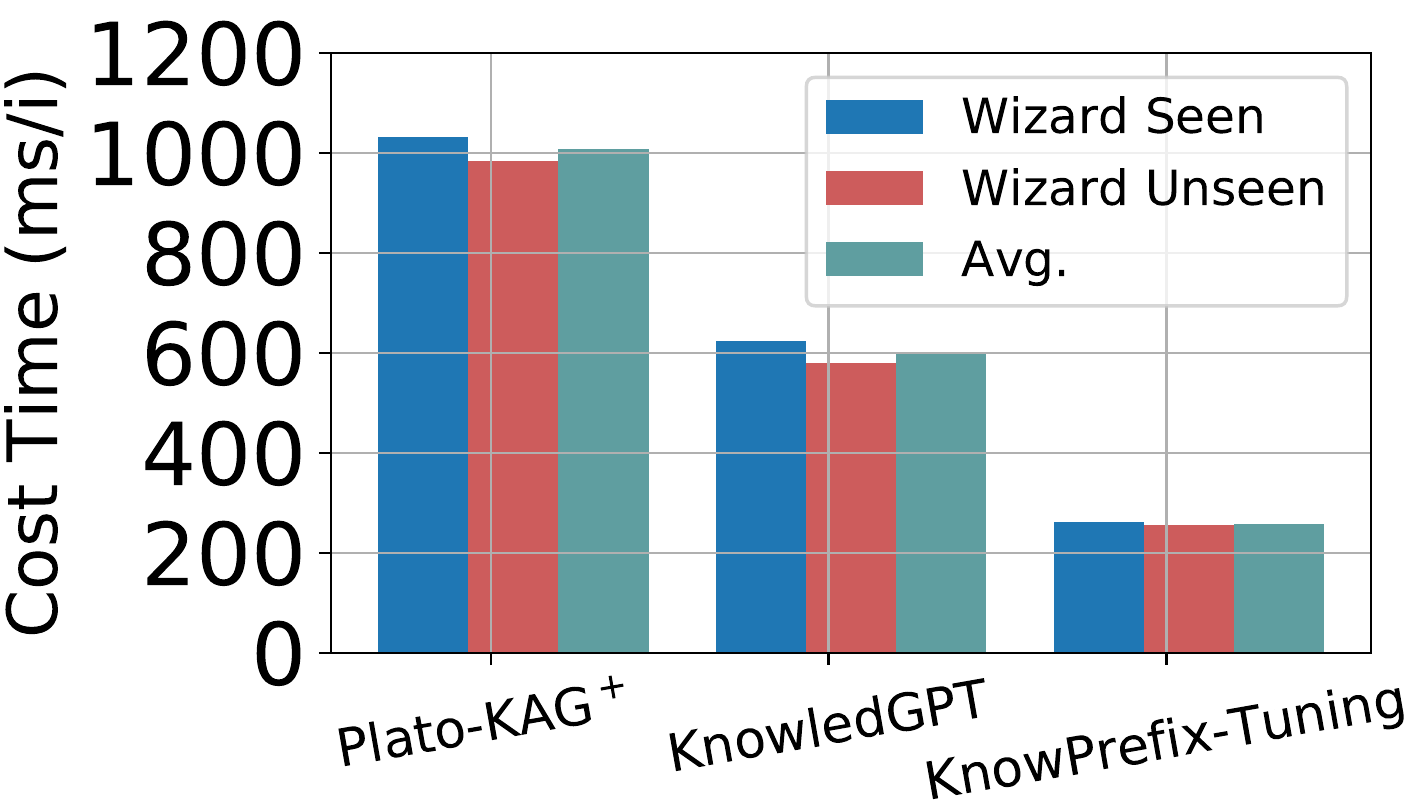}
\caption{The inference time comparison across different baselines. ``ms/i'' indicates millisecond per instance.}
\label{cost_time}
\end{figure}

\begin{table*}[t]
\centering
\caption{Ablation study on Wizard and CMU\_DoG datasets. ``-'' means removing the corresponding part.}
\renewcommand\arraystretch{1.1}
\resizebox{0.85\linewidth}{!}{
\begin{tabular}{l ccc ccc ccc}
    \hline
    \hline
    
    \multirow{3}*{\textbf{Models}} & 
    \multicolumn{3}{c}{\textbf{Wizard Seen}} & 
    \multicolumn{3}{c}{\textbf{Wizard Unseen}} & 
    \multicolumn{3}{c}{\textbf{CMU\_DoG}} \cr

    \cmidrule(lr){2-4} \cmidrule(lr){5-7} \cmidrule(lr){8-10} &
    
    \textbf{PPL}$\downarrow$ & \textbf{F1}$\uparrow$ & \textbf{KF1}$\uparrow$ & 
    \textbf{PPL}$\downarrow$ & \textbf{F1}$\uparrow$ & \textbf{KF1}$\uparrow$ & 
    \textbf{PPL}$\downarrow$ & \textbf{F1}$\uparrow$ & \textbf{KF1}$\uparrow$ \\

    \hline

    KnowPrefix-Tuning (BART{\footnotesize{LARGE}}) & \textbf{14.0} & \textbf{20.3} & \textbf{17.4} & \textbf{17.5} & \textbf{18.3} & \textbf{14.9} & \textbf{16.3} & \textbf{14.6} & \textbf{5.2} \\

    \quad - Interactive Re-parameterization & 14.8 & 19.3 & 16.8 & 18.5 & 17.4 & 13.7 & 18.7 & 13.7 & 4.5 \\

    \quad - Stage-I & 16.0 & 18.7 & 16.3 & 20.2 & 16.4 & 13.0 & 20.4 & 13.3 & 3.8 \\
    
    

    \hline

    KnowPrefix-Tuning (GPT2{\footnotesize{LARGE}}) & 15.2 & \textbf{20.1} & \textbf{18.0} & \textbf{19.3} & \textbf{18.0} & \textbf{14.6} & \textbf{17.4} & \textbf{14.1} & \textbf{5.3} \\

    \quad - Interactive Re-parameterization & 15.7 & 19.2 & 17.2 & 20.5 & 17.2 & 13.8 & 19.7 & 13.2 & 4.5 \\

    \quad - Stage-I & 17.8 & 18.5 & 16.7 & 22.0 & 16.3 & 12.8 & 20.9 & 12.8 & 3.9 \\
    
    
    
    \hline
    \hline
\end{tabular}
}
\label{ablation_results}
\end{table*}

\subsubsection{Ablation Study}
To facilitate the study of how each component of the model influence the overall performance, we conduct the ablation study for both BART{\footnotesize{LARGE}} and GPT2{\footnotesize{LARGE}} on Wizard and CMU\_DoG datasets.
We compare the proposed KnowPrefix-Tuning with the following variants: 
(1) \emph{- Interactive Re-parameterization}: 
The interactive re-parameterization module is replaced by the re-parameterization module as we defined in equation \ref{prefix_repar}.  
(2) \emph{- Stage-I}: 
The stage-I is removed during the knowledgeable response generation. 
Thus, the model generates responses without considering the grounded knowledge (i.e., the knowledge labeled by human annotators.).
For a fair comparison, we uniform the number of parameters used in this variant and the KnowPrefix-Tuning by increasing the variant's prefix length to 40.
Table \ref{ablation_results} presents the results, we can conclude that:
(1) Without the Interactive Re-parameterization module, the performance of both BART{\footnotesize{LARGE}} and GPT2{\footnotesize{LARGE}} drops significantly.
We believe that the interactive re-parameterization module allows the interaction between the embedding of prefixes and PLM, which is helpful to knowledgeable response generation.
(2) Without Stage-I, the performance of both BART{\footnotesize{LARGE}} and GPT2{\footnotesize{LARGE}} goes down.
It indicates the effectiveness of the proposed two-stage framework.
Even without equipping with the proposed interactive re-parameterization mechanism, the proposed two-stage framework incorporates the context-relevant knowledge into the generated response and substantially improves the generation of knowledgeable responses.



\subsubsection{Inference Time Efficiency}

To verify the inference time efficiency of our proposed method,  we compare the proposed KnowPrefix-Tuning (BART-LARGE) with two strong retrieval-based methods, which are PLATO-KAG$^+$ and KnowledGPT, under the same inference implementation setting for a fair comparison. 
Figure \ref{cost_time} reports the results on Wizard Seen and Unseen test set.
We observe that  KnowPrefix-Tuning is around $3\times$ faster and $5\times$ faster than KnowledGPT and PLATO-KAG$^+$ during the inference stage, respectively. 
We believe it is because that the proposed KnowPrefix-Tuning bypasses the retrieval process and doesn't require augmenting the retrieved knowledge with the input dialogue context, saving much time during inference.

\subsubsection{Impact of Pretrained Language Models}

\begin{figure}[t]
\centering
    \subfigure[Knowledge Generation]
    {
    \begin{minipage}{5.cm}
    \label{know_gen_study}
    \centering
    \includegraphics[width=5.cm]{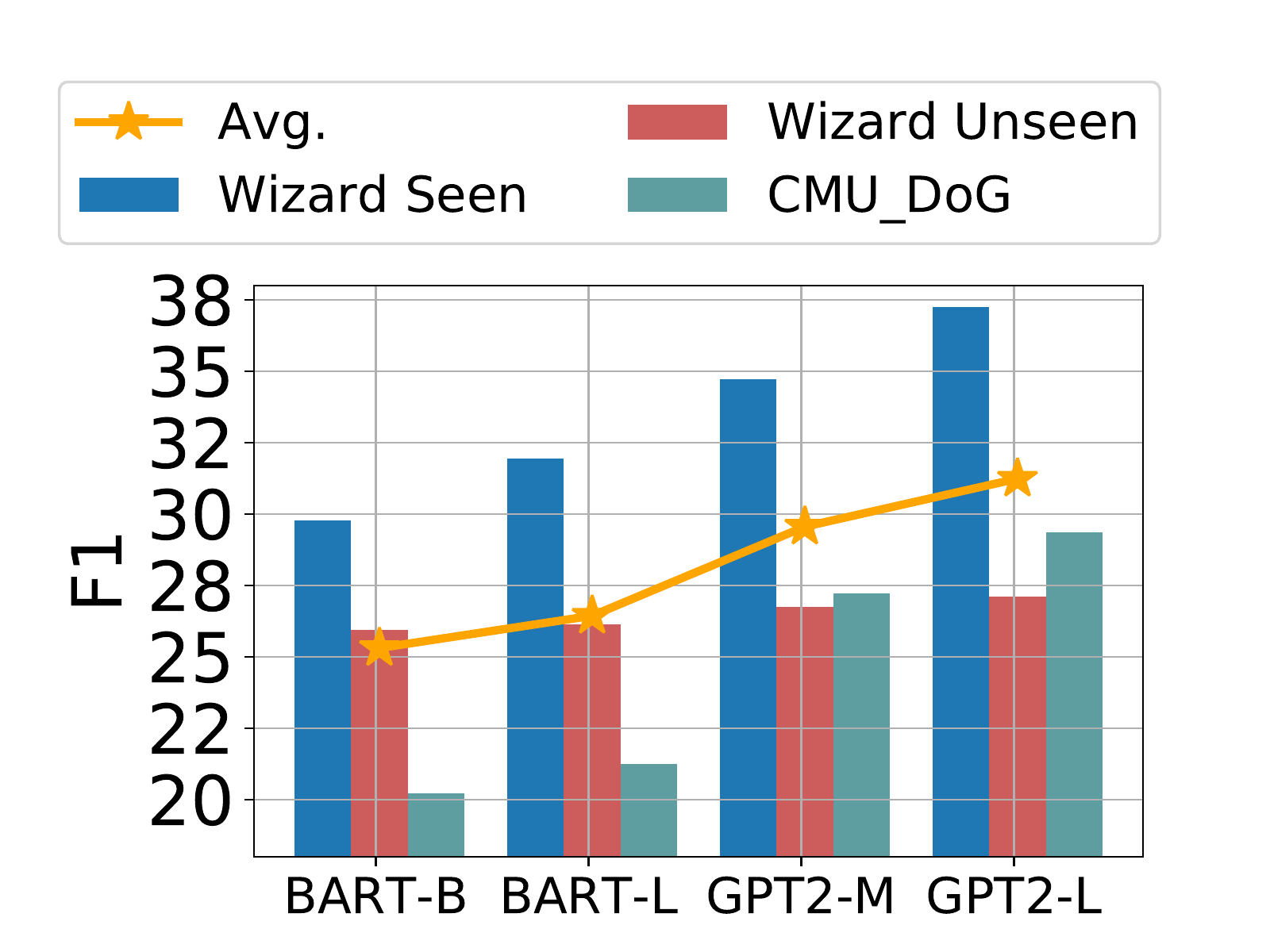}
    \end{minipage}
    }
    \subfigure[Response Generation]
    {
    \begin{minipage}{5.cm}
    \label{resp_gen_study}
    \centering
    \includegraphics[width=5.cm]{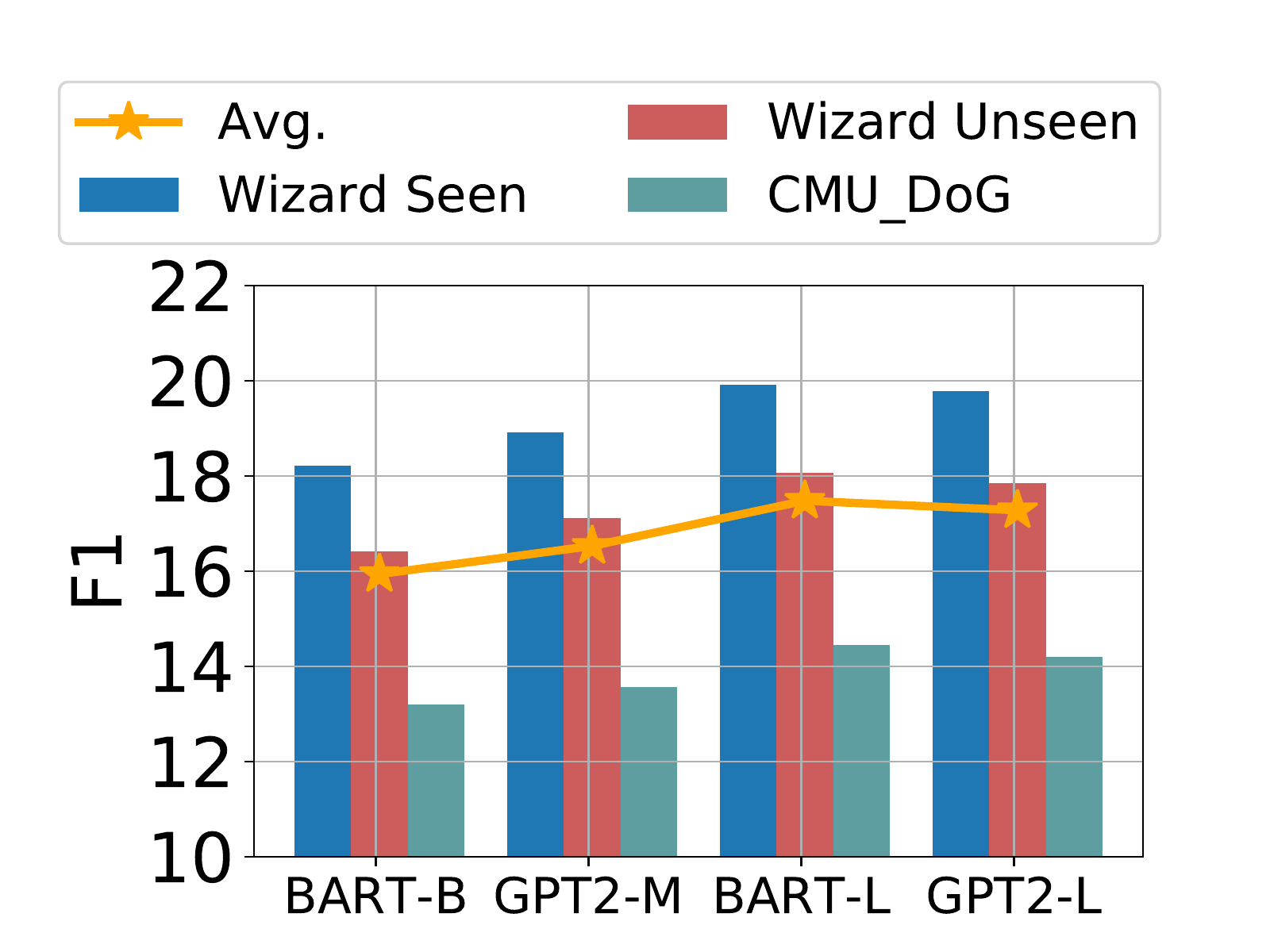}
    \end{minipage}
    }
\caption{Impact of PLMs. ``BART-B'', ``BART-L'', ``GPT2-M'' and ``GPT2-L'' denote the BART-BASE, BART-LARGE, GPT2-MEDIUM and GPT2-LARGE, respectively. ``Avg'' is the average F1 score across the three datasets.} 
\label{model_size_impact}
\end{figure}

To further study how does the size of PLMs impacts the performance of the KnowPrefix-Tuning. 
We additionally employ BART{\footnotesize{BASE}} (139M) and GPT2{\footnotesize{MEDIUM}} (345M) to realize the knowledge generation and the response generation.
Note that our approach does not require to explicitly augment the generated knowledge with the dialogue context to generate responses.
We evaluate the quality of the generated knowledge to investigate whether there is a connection between the generated knowledge and the generated responses when using the proposed KnowPrefix-tuning approach.

Figure \ref{model_size_impact} presents the results. 
We have the following obeservations:
(1) For knowledge generation, the larger model generates better knowledge, while it is not significant on the Wizard Unseen dataset.
We believe that it is still a challenge for the large language model to generate better knowledge with unseen topics. 
(2) For knowledgeable response generation, the quality of the generated responses improves with the model size boost. 
In addition, we observe that the model which generates better knowledge can also generate better responses with the same backbone PLMs.
It indicates that the proposed approach equipped with well-learned knowledge prefix can elicit proper knowledge from PLMs to guide the knowledgeable response generation.


\subsubsection{Impact of Prefix Length}

We investigate whether Prefix-Tuning can achieve better performance than KnowPrefix-Tuning by increasing its prefix length.
We employ BART\texttt{LARGE} as the backbone model for each variant.
Table \ref{prefix_length_impact} shows the results. 
We observe that increasing prefix length will introduce more parameters.
However, the performance of Prefix-Tuning seems not to be better even if it introduce additional parameters comparable to KnowPrefix-Tuning.
We think that although the longer prefix length introduces more trainable parameters and increases the generalization ability of the model, too long a prefix length may lead the model tend to overfit the training dataset, degrading the knowledgeable response generation ability.

\begin{table}[t]
\centering
\caption{Comparison of KnowPrefix-Tuning and Prefix-Tuning with different prefix length ``$l$''. We employ BART\texttt{LARGE} as the backbone model for each variant.}
\renewcommand\arraystretch{1.1}
\resizebox{0.75\linewidth}{!}{
\begin{tabular}{l ccc ccc c}
    \hline
    \hline
    
    \multirow{3}*{\textbf{Models}} & 
    \multicolumn{3}{c}{\textbf{Wizard Seen}} & 
    \multicolumn{3}{c}{\textbf{Wizard Unseen}} & 
    \multirow{3}*{\textbf{\#PSD}} \cr

    \cmidrule(lr){2-4} \cmidrule(lr){5-7} &
    
    \textbf{PPL$\downarrow$} & \textbf{F1$\uparrow$} & \textbf{KF1$\uparrow$} & 
    \textbf{PPL$\downarrow$} & \textbf{F1$\uparrow$} & \textbf{KF1$\uparrow$} \\

    \hline

    KnowPrefix-Tuning ($l=10$) & 14.4 & 19.8 & 17.0 & 18.2 & 17.8 & 14.4 & $1.5\times 10^6$ \\

    KnowPrefix-Tuning ($l=20$) & 14.0 & 20.3 & 17.4 & 17.5 & 18.3 & 14.9 & $2.9\times 10^6$ \\

    Prefix-Tuning ($l=20$) & 15.2 & 19.2 & 16.5 & 19.4 & 16.9 & 13.4 & $1.5\times 10^6$\\

    Prefix-Tuning ($l=40$) & 16.0 & 18.7 & 16.3 & 20.2 & 16.4 & 13.0 & $3.0 \times 10^6$\\

    Prefix-Tuning ($l=60$) & 15.9 & 18.5 & 16.4 & 20.5 & 16.1 & 13.3 & $4.4 \times 10^6$\\
    
    \hline
    \hline
\end{tabular}
}
\label{prefix_length_impact}
\end{table}

\subsubsection{Case Study}
Table \ref{wizard_case_1} and Table \ref{wizard_case_2} show the cases from the Wizard test set, from which we observe that the generated responses of the KnowPrefix-Tuning not only leverage the knowledge relevant to the dialogue context but also ensure the correctness of the utilized knowledge.
As shown in Table \ref{wizard_case_1}, the responses generated by KnowPrefix-Tuning on both Test Seen and Test Unseen contain essential facts that appeared in the golden answer.
We believe that the learned knowledge prefix effectively provides facts relevant to the dialogue context, which is indispensable to guide knowledgeable response generation.
From Table \ref{wizard_case_2}, we observe that the responses generated by KnowPrefix-Tuning contain factually correct knowledge even if the corresponding evidence is not explicitly provided.
We suspect that it is because the parameters of the language model are frozen during the fine-tuning procedure. 
Thus, the knowledge inherently encoded in the pre-trained language model is not disturbed, which can be prompted to provide evidence for knowledgeable response generation.

\begin{table*}[]
\caption{Cases from the test seen and test unseen data of Wizard. The underline text indicates the essential facts that appeared in the golden response.}
\centering
\setlength{\tabcolsep}{3pt}
\renewcommand\arraystretch{1.2}
\resizebox{1.\linewidth}{!}{
\begin{tabular}{rp{140pt}|rp{140pt}}
    \hline
    \hline
    \multicolumn{4}{c}{\textbf{Dialogue Context}} \\
    
    \hline
    
    \multicolumn{2}{c|}{\textbf{Test Seen}} &
    \multicolumn{2}{c}{\textbf{Test Unseen}} \\
    
    \hline
    
    \emph{Wizard:} & 
    Well the rights to this forumula was obtained by the TIP corporation. 
    &
    \emph{Wizard:} & 
    I enjoy hunting. This refers to the killing or trapping animals, or pursuing or tracking them. \\
    
    \hline

    \emph{Apprentice:} & 
    Do you know how they came up with the other flavors?  &
    \emph{Apprentice:} & 
    How long have you been hunting? \\
    
    \hline
    \multicolumn{4}{c}{\textbf{Truth Response}}  \\
    \hline
    
    \emph{Truth:} & 
    Well the diet verstion was created in 1988, followed by mountain dew red which was discontinued in the same year. & 
    \emph{Truth:} & 
    20 years now! Excessive hunting has also heavily contributed to the endangerment, extirpation and extinction of many animals. \\

    \hline
    \multicolumn{4}{c}{\textbf{Generated Response}}  \\
    \hline
    

    KnowExpert$_{\mathrm{w}}$: &
    Well they were created by a group of volunteers.&
    KnowExpert$_{\mathrm{w}}$: &
    I have been hunting for about 10 years. \\

    \hline
    Prefix-Tuning: &
    Well it was first introduced in the United States in 1953 as a soft drink. &
    Prefix-Tuning: &
    I have been hunting since i was a child. It is the practice of hunting animals for food or raw materials \\

    \hline
    Fine-Tuning: &
    I don't but i know that it was introduced in the US in 1964. & 
    Fine-Tuning: &
    I have been hunting for about \underline{20 years}. It is a very difficult activity. \\

    \hline
    KnowPrefix-Tuning: &
    Well the first \underline{mountain dew} was \underline{created in} 1953 by a pharmacist named Charles Alderton. &
    KnowPrefix-Tuning: &
    I have been hunting for about \underline{20 years}. I like to hunt for food, sport, and for medicinal purposes. \\

    \hline
    \hline

\end{tabular}
}
\label{wizard_case_1}
\end{table*}

\begin{table*}[]
\centering
\caption{Cases from the test seen and test unseen data of Wizard. The facts with boldface have been verified on the Internet.}
\setlength{\tabcolsep}{3pt}
\renewcommand\arraystretch{1.2}
\resizebox{1.\linewidth}{!}{
\begin{tabular}{rp{140pt}|rp{140pt}}
    \hline
    \hline
    \multicolumn{4}{c}{\textbf{Dialogue Context}} \\
    
    \hline
    
    \multicolumn{2}{c|}{\textbf{Test Seen}} &
    \multicolumn{2}{c}{\textbf{Test Unseen}} \\
    
    \hline
    
    \emph{Wizard:} & 
    Hey, what do you think of Mcdonald's dollar menu cheeseburgers?
    &
    \emph{Wizard:} & 
    I love American football also know as gridiron. \\
    
    \hline

    \emph{Apprentice:} & 
    I think it would be worth the money spent for trying them. &
    \emph{Apprentice:} & 
    It's a pretty fun sport to watch, also I have never heard it be called that! \\
    
    \hline
    \multicolumn{4}{c}{\textbf{Truth Response}}  \\
    \hline
    
    \emph{Truth:} & 
    It is actually. I don't know about the breakfast. & 
    \emph{Truth:} & 
    Yes, it originated from association football and rugby. \\

    \hline
    \multicolumn{4}{c}{\textbf{Generated Response}}  \\
    \hline
    

    KnowExpert$_{\mathrm{w}}$: &
    I agree, I love the fact that they have a menu of different items. &
    KnowExpert$_{\mathrm{w}}$: &
    It is played on a rectangular field with a goal at each end. \\

    \hline
    Prefix-Tuning: &
    Yes, I agree. I also like their chicken nuggets and french fries. &
    Prefix-Tuning: &
    It is called gridiron football in the USA. It is a team sport played on a rectangular field.\\

    \hline
    Finetune: &
    I agree, i love the cheeseburgers. They're so good! Mcdonald's also owns and operates Sam's club. &
    Finetune: &
    It is played on a rectangular field with a goal at each end. It is also called gridiron football. \\

    \hline
    KnowPrefix-Tuning: &
    I agree, they are so good. \textbf{They are a hamburger stand that was founded in 1940.} &
    KnowPrefix-Tuning: &
    \textbf{It is played by 250 million people in over 200 countries}. It is the most popular sport in the world. \\

    \hline
    \hline
\end{tabular}
}
\label{wizard_case_2}
\end{table*}
\section{Conclusions}
In this paper, we propose KnowPrefix-Tuning to handle the knowledge-grounded dialogue generation task. 
The proposed method bypasses the retrieval process and does not require fine-tuning the entire PLM. 
In addition, the proposed interactive re-parameterization mechanism allows the interaction between the embedding of prefixes and PLM.
Experiments on two commonly-used knowledge-grounded dialogue datasets demonstrate the effectiveness of our approach.

\subsubsection{Acknowledgments.}

We thank all the anonymous reviewers for their insightful comments. This work was supported in part by the National Natural Science Foundation of China (Grant Nos. 62276017, U1636211, 61672081), the 2022 Tencent Big Travel Rhino-Bird Special Research Program, and the Fund of the State Key Laboratory of Software Development Environment (Grant No. SKLSDE-2021ZX-18).

\bibliographystyle{splncs04}
\bibliography{anthology}

\begin{thebibliography}{10}
\providecommand{\url}[1]{\texttt{#1}}
\providecommand{\urlprefix}{URL }
\providecommand{\doi}[1]{https://doi.org/#1}

\bibitem{bao-etal-2021-plato}
Bao, S., He, H., Wang, F., Wu, H., Wang, H., Wu, W., Guo, Z., Liu, Z., Xu, X.:
  {PLATO-2}: Towards building an open-domain chatbot via curriculum learning.
  In: Findings of the Association for Computational Linguistics: ACL-IJCNLP
  2021. pp. 2513--2525. Association for Computational Linguistics, Online (Aug
  2021). \doi{10.18653/v1/2021.findings-acl.222}

\bibitem{brown2020language}
Brown, T., Mann, B., Ryder, N., Subbiah, M., Kaplan, J.D., Dhariwal, P.,
  Neelakantan, A., Shyam, P., Sastry, G., Askell, A., et~al.: Language models
  are few-shot learners. Advances in neural information processing systems
  \textbf{33},  1877--1901 (2020)

\bibitem{cui2021commonsense}
Cui, L., Cheng, S., Wu, Y., Zhang, Y.: On commonsense cues in bert for solving
  commonsense tasks. In: Findings of the Association for Computational
  Linguistics: ACL-IJCNLP 2021. pp. 683--693 (2021)

\bibitem{cui2021knowledge}
Cui, L., Wu, Y., Liu, S., Zhang, Y.: Knowledge enhanced fine-tuning for better
  handling unseen entities in dialogue generation. In: Proceedings of the 2021
  Conference on Empirical Methods in Natural Language Processing. pp.
  2328--2337 (2021)

\bibitem{davison2019commonsense}
Davison, J., Feldman, J., Rush, A.M.: Commonsense knowledge mining from
  pretrained models. In: Proceedings of the 2019 conference on empirical
  methods in natural language processing and the 9th international joint
  conference on natural language processing (EMNLP-IJCNLP). pp. 1173--1178
  (2019)

\bibitem{dinan2018wizard}
Dinan, E., Roller, S., Shuster, K., Fan, A., Auli, M., Weston, J.: Wizard of
  wikipedia: Knowledge-powered conversational agents. In: International
  Conference on Learning Representations (2018)

\bibitem{fleiss1971measuring}
Fleiss, J.L.: Measuring nominal scale agreement among many raters.
  Psychological bulletin  \textbf{76}(5), ~378 (1971)

\bibitem{ghazvininejad2018knowledge}
Ghazvininejad, M., Brockett, C., Chang, M.W., Dolan, B., Gao, J., Yih, W.t.,
  Galley, M.: A knowledge-grounded neural conversation model. In: Proceedings
  of the AAAI Conference on Artificial Intelligence. vol.~32 (2018)

\bibitem{guu2020retrieval}
Guu, K., Lee, K., Tung, Z., Pasupat, P., Chang, M.: Retrieval augmented
  language model pre-training. In: International Conference on Machine
  Learning. pp. 3929--3938. PMLR (2020)

\bibitem{holtzman2019curious}
Holtzman, A., Buys, J., Du, L., Forbes, M., Choi, Y.: The curious case of
  neural text degeneration. In: International Conference on Learning
  Representations (2019)

\bibitem{huang2021plato}
Huang, X., He, H., Bao, S., Wang, F., Wu, H., Wang, H.: Plato-kag: Unsupervised
  knowledge-grounded conversation via joint modeling. In: Proceedings of the
  3rd Workshop on Natural Language Processing for Conversational AI. pp.
  143--154 (2021)

\bibitem{karimi-mahabadi-etal-2022-prompt}
Karimi~Mahabadi, R., Zettlemoyer, L., Henderson, J., Mathias, L., Saeidi, M.,
  Stoyanov, V., Yazdani, M.: Prompt-free and efficient few-shot learning with
  language models. In: Proceedings of the 60th Annual Meeting of the
  Association for Computational Linguistics (Volume 1: Long Papers). pp.
  3638--3652. Association for Computational Linguistics, Dublin, Ireland (May
  2022). \doi{10.18653/v1/2022.acl-long.254}

\bibitem{kenton2019bert}
Kenton, J.D.M.W.C., Toutanova, L.K.: Bert: Pre-training of deep bidirectional
  transformers for language understanding. In: Proceedings of NAACL-HLT. pp.
  4171--4186 (2019)

\bibitem{kim2019sequential}
Kim, B., Ahn, J., Kim, G.: Sequential latent knowledge selection for
  knowledge-grounded dialogue. In: International Conference on Learning
  Representations (2019)

\bibitem{komeili2022internet}
Komeili, M., Shuster, K., Weston, J.: Internet-augmented dialogue generation.
  In: Proceedings of the 60th Annual Meeting of the Association for
  Computational Linguistics (Volume 1: Long Papers). pp. 8460--8478 (2022)

\bibitem{lester2021power}
Lester, B., Al-Rfou, R., Constant, N.: The power of scale for
  parameter-efficient prompt tuning. In: Proceedings of the 2021 Conference on
  Empirical Methods in Natural Language Processing. pp. 3045--3059 (2021)

\bibitem{lewis2020bart}
Lewis, M., Liu, Y., Goyal, N., Ghazvininejad, M., Mohamed, A., Levy, O.,
  Stoyanov, V., Zettlemoyer, L.: Bart: Denoising sequence-to-sequence
  pre-training for natural language generation, translation, and comprehension.
  In: Proceedings of the 58th Annual Meeting of the Association for
  Computational Linguistics. pp. 7871--7880 (2020)

\bibitem{lewis2020retrieval}
Lewis, P., Perez, E., Piktus, A., Petroni, F., Karpukhin, V., Goyal, N.,
  K{\"u}ttler, H., Lewis, M., Yih, W.t., Rockt{\"a}schel, T., et~al.:
  Retrieval-augmented generation for knowledge-intensive nlp tasks. Advances in
  Neural Information Processing Systems  \textbf{33},  9459--9474 (2020)

\bibitem{li2020zero}
Li, L., Xu, C., Wu, W., Zhao, Y., Zhao, X., Tao, C.: Zero-resource
  knowledge-grounded dialogue generation. Advances in Neural Information
  Processing Systems  \textbf{33},  8475--8485 (2020)

\bibitem{li-liang-2021-prefix}
Li, X.L., Liang, P.: Prefix-tuning: Optimizing continuous prompts for
  generation. In: Proceedings of the 59th Annual Meeting of the Association for
  Computational Linguistics and the 11th International Joint Conference on
  Natural Language Processing (Volume 1: Long Papers). pp. 4582--4597.
  Association for Computational Linguistics, Online (Aug 2021).
  \doi{10.18653/v1/2021.acl-long.353}

\bibitem{li2022eliciting}
Li, Y., Zhao, J., Lyu, M.R., Wang, L.: Eliciting knowledge from large
  pre-trained models for unsupervised knowledge-grounded conversation. arXiv
  preprint arXiv:2211.01587  (2022)

\bibitem{li-etal-2022-knowledge}
Li, Y., Peng, B., Shen, Y., Mao, Y., Liden, L., Yu, Z., Gao, J.:
  Knowledge-grounded dialogue generation with a unified knowledge
  representation. In: Proceedings of the 2022 Conference of the North American
  Chapter of the Association for Computational Linguistics: Human Language
  Technologies. pp. 206--218. Association for Computational Linguistics,
  Seattle, United States (Jul 2022). \doi{10.18653/v1/2022.naacl-main.15}

\bibitem{lian2019learning}
Lian, R., Xie, M., Wang, F., Peng, J., Wu, H.: Learning to select knowledge for
  response generation in dialog systems. In: IJCAI International Joint
  Conference on Artificial Intelligence. p.~5081 (2019)

\bibitem{mbart}
Liu, Y., Gu, J., Goyal, N., Li, X., Edunov, S., Ghazvininejad, M., Lewis, M.,
  Zettlemoyer, L.: Multilingual denoising pre-training for neural machine
  translation. Trans. Assoc. Comput. Linguistics  \textbf{8},  726--742 (2020).
  \doi{10.1162/tacl\_a\_00343}

\bibitem{liu2022multi}
Liu, Z., Patwary, M., Prenger, R., Prabhumoye, S., Ping, W., Shoeybi, M.,
  Catanzaro, B.: Multi-stage prompting for knowledgeable dialogue generation.
  In: Findings of the Association for Computational Linguistics: ACL 2022. pp.
  1317--1337 (2022)

\bibitem{loshchilov2018fixing}
Loshchilov, I., Hutter, F.: Fixing weight decay regularization in adam  (2018)

\bibitem{meng2020dukenet}
Meng, C., Ren, P., Chen, Z., Sun, W., Ren, Z., Tu, Z., Rijke, M.d.: Dukenet: A
  dual knowledge interaction network for knowledge-grounded conversation. In:
  Proceedings of the 43rd International ACM SIGIR Conference on Research and
  Development in Information Retrieval. pp. 1151--1160 (2020)

\bibitem{petroni2019language}
Petroni, F., Rockt{\"a}schel, T., Riedel, S., Lewis, P., Bakhtin, A., Wu, Y.,
  Miller, A.: Language models as knowledge bases? In: Proceedings of the 2019
  Conference on Empirical Methods in Natural Language Processing and the 9th
  International Joint Conference on Natural Language Processing (EMNLP-IJCNLP).
  pp. 2463--2473 (2019)

\bibitem{prabhumoye2021focused}
Prabhumoye, S., Hashimoto, K., Zhou, Y., Black, A.W., Salakhutdinov, R.:
  Focused attention improves document-grounded generation. In: Proceedings of
  the 2021 Conference of the North American Chapter of the Association for
  Computational Linguistics: Human Language Technologies. pp. 4274--4287 (2021)

\bibitem{radford2019language}
Radford, A., Wu, J., Child, R., Luan, D., Amodei, D., Sutskever, I., et~al.:
  Language models are unsupervised multitask learners. OpenAI Blog  (2019)

\bibitem{roller2021recipes}
Roller, S., Dinan, E., Goyal, N., Ju, D., Williamson, M., Liu, Y., Xu, J., Ott,
  M., Smith, E.M., Boureau, Y.L., et~al.: Recipes for building an open-domain
  chatbot. In: Proceedings of the 16th Conference of the European Chapter of
  the Association for Computational Linguistics: Main Volume. pp. 300--325
  (2021)

\bibitem{schick2021s}
Schick, T., Sch{\"u}tze, H.: It’s not just size that matters: Small language
  models are also few-shot learners. In: Proceedings of the 2021 Conference of
  the North American Chapter of the Association for Computational Linguistics:
  Human Language Technologies. pp. 2339--2352 (2021)

\bibitem{shin2020autoprompt}
Shin, T., Razeghi, Y., Logan~IV, R.L., Wallace, E., Singh, S.: Autoprompt:
  Eliciting knowledge from language models with automatically generated
  prompts. In: Proceedings of the 2020 Conference on Empirical Methods in
  Natural Language Processing (EMNLP). pp. 4222--4235 (2020)

\bibitem{shuster2021retrieval}
Shuster, K., Poff, S., Chen, M., Kiela, D., Weston, J.: Retrieval augmentation
  reduces hallucination in conversation. In: Findings of the Association for
  Computational Linguistics: EMNLP 2021. pp. 3784--3803 (2021)

\bibitem{sun-etal-2022-multimodal}
Sun, Q., Wang, Y., Xu, C., Zheng, K., Yang, Y., Hu, H., Xu, F., Zhang, J.,
  Geng, X., Jiang, D.: Multimodal dialogue response generation. In: Proceedings
  of the 60th Annual Meeting of the Association for Computational Linguistics
  (Volume 1: Long Papers). pp. 2854--2866. Association for Computational
  Linguistics, Dublin, Ireland (May 2022). \doi{10.18653/v1/2022.acl-long.204}

\bibitem{sun-etal-2022-stylized}
Sun, Q., Xu, C., Hu, H., Wang, Y., Miao, J., Geng, X., Chen, Y., Xu, F., Jiang,
  D.: Stylized knowledge-grounded dialogue generation via disentangled template
  rewriting. In: Proceedings of the 2022 Conference of the North American
  Chapter of the Association for Computational Linguistics: Human Language
  Technologies. pp. 3304--3318. Association for Computational Linguistics,
  Seattle, United States (Jul 2022). \doi{10.18653/v1/2022.naacl-main.241}

\bibitem{DBLP:journals/corr/abs-2212-10400}
Sun, W., Shi, Z., Gao, S., Ren, P., de~Rijke, M., Ren, Z.: Contrastive learning
  reduces hallucination in conversations. CoRR  \textbf{abs/2212.10400} (2022).
  \doi{10.48550/arXiv.2212.10400}

\bibitem{vaswani2017attention}
Vaswani, A., Shazeer, N., Parmar, N., Uszkoreit, J., Jones, L., Gomez, A.N.,
  Kaiser, {\L}., Polosukhin, I.: Attention is all you need. Advances in neural
  information processing systems  \textbf{30} (2017)

\bibitem{wang2021can}
Wang, C., Liu, P., Zhang, Y.: Can generative pre-trained language models serve
  as knowledge bases for closed-book qa? In: Proceedings of the 59th Annual
  Meeting of the Association for Computational Linguistics and the 11th
  International Joint Conference on Natural Language Processing (Volume 1: Long
  Papers). pp. 3241--3251 (2021)

\bibitem{xu2022retrieval}
Xu, Y., Ishii, E., Cahyawijaya, S., Liu, Z., Winata, G.I., Madotto, A., Su, D.,
  Fung, P.: Retrieval-free knowledge-grounded dialogue response generation with
  adapters. In: Proceedings of the Second DialDoc Workshop on Document-grounded
  Dialogue and Conversational Question Answering. pp. 93--107 (2022)

\bibitem{zhao-etal-2022-domain}
Zhao, L., Zheng, F., Zeng, W., He, K., Xu, W., Jiang, H., Wu, W., Wu, Y.:
  Domain-oriented prefix-tuning: Towards efficient and generalizable
  fine-tuning for zero-shot dialogue summarization. In: Proceedings of the 2022
  Conference of the North American Chapter of the Association for Computational
  Linguistics: Human Language Technologies. pp. 4848--4862. Association for
  Computational Linguistics, Seattle, United States (Jul 2022).
  \doi{10.18653/v1/2022.naacl-main.357}

\bibitem{zhao2017learning}
Zhao, T., Zhao, R., Eskenazi, M.: Learning discourse-level diversity for neural
  dialog models using conditional variational autoencoders. In: Proceedings of
  the 55th Annual Meeting of the Association for Computational Linguistics
  (Volume 1: Long Papers). pp. 654--664 (2017)

\bibitem{zhao-etal-2022-learning-express}
Zhao, X., Fu, T., Tao, C., Wu, W., Zhao, D., Yan, R.: Learning to express in
  knowledge-grounded conversation. In: Proceedings of the 2022 Conference of
  the North American Chapter of the Association for Computational Linguistics:
  Human Language Technologies. pp. 2258--2273. Association for Computational
  Linguistics, Seattle, United States (Jul 2022).
  \doi{10.18653/v1/2022.naacl-main.164}

\bibitem{zhao2020knowledge}
Zhao, X., Wu, W., Xu, C., Tao, C., Zhao, D., Yan, R.: Knowledge-grounded
  dialogue generation with pre-trained language models. In: Proceedings of the
  2020 Conference on Empirical Methods in Natural Language Processing (EMNLP).
  pp. 3377--3390 (2020)

\bibitem{zhou2018dataset}
Zhou, K., Prabhumoye, S., Black, A.W.: A dataset for document grounded
  conversations. In: Proceedings of the 2018 Conference on Empirical Methods in
  Natural Language Processing. pp. 708--713 (2018)

\bibitem{zhou2020evaluating}
Zhou, X., Zhang, Y., Cui, L., Huang, D.: Evaluating commonsense in pre-trained
  language models. In: Proceedings of the AAAI Conference on Artificial
  Intelligence. vol.~34, pp. 9733--9740 (2020)

\end{thebibliography}

\end{sloppypar}
\end{document}